\documentclass[lettersize,journal]{IEEEtran}
\usepackage{amsmath,amsfonts}
\usepackage{algorithm}
\usepackage{array}
\usepackage{bm}
\usepackage[caption=false,font=normalsize,labelfont=sf,textfont=sf]{subfig}
\usepackage{textcomp}
\usepackage{stfloats}
\usepackage{url}
\usepackage{verbatim}
\usepackage{graphicx}
\usepackage{cite}
\usepackage{amsmath}
\usepackage{algpseudocode}
\begin{document}

\title{DRUformer: Enhancing the driving scene important object detection with driving relationship self-understanding}

\author{Yingjie Niu,~\IEEEmembership{Student Member,~IEEE,}
        Ming Ding,~\IEEEmembership{Member,~IEEE,}
        Keisuke Fujii,~\IEEEmembership{Member,~IEEE,}
        Kento Ohtani,~\IEEEmembership{Member,~IEEE,}
        Alexander Carballo,~\IEEEmembership{Member,~IEEE,}
        Kazuya Takeda,~\IEEEmembership{Senior Member,~IEEE}
        % <-this % stops a space
\thanks{Mr. Niu is with the Department of Informatics, Nagoya University, Japan. e-mail: niu.yingjie@g.sp.m.is.nagoya-u.ac.jp}}% <-this % stops a space

\markboth{Journal of \LaTeX\ Class Files,~Vol.~14, No.~8, August~2021}%
{Shell \MakeLowercase{\textit{et al.}}: A Sample Article Using IEEEtran.cls for IEEE Journals}

\IEEEpubid{0000--0000/00\$00.00~\copyright~2021 IEEE}

\maketitle

\begin{abstract}
Traffic accidents frequently lead to fatal injuries, contributing to over 50 million deaths until 2023. 
To mitigate driving hazards and ensure personal safety, it is crucial to assist vehicles in anticipating important objects during travel.
Previous research on important object detection primarily assessed the importance of individual participants, treating them as independent entities and frequently neglecting the interconnections among these participants. 
Unfortunately, this approach has proven less effective in detecting important objects in complex scenarios.
In this work, we introduce Driving scene Relationship self-Understanding transformer (DRUformer), designed to enhance the important object detection task. 
The DRUformer is a transformer-based multi-modal important object detection model that takes into account the relationships between all the participants in the driving scenario. 
Recognizing that driving intention also significantly affects the detection of important objects during driving, we have incorporated a module for embedding driving intention.
To assess the performance of our approach, we conducted comparative experiments on the DRAMA dataset, pitting our model against other state-of-the-art (SOTA) models. 
The results demonstrated a noteworthy 16.2\% improvement in mIoU and a substantial 12.3\% boost in ACC compared to SOTA methods.
Furthermore, we conducted a qualitative analysis of our model's ability to detect important objects across different road scenarios and classes, highlighting its effectiveness in diverse contexts. 
Finally, we conducted various ablation studies to assess the efficiency of the proposed modules in our DRUformer model.

The code will be made publicly available on this link upon acceptance of the paper. 
\url{https://github.com/oniu-uin0/DRU_former}
\end{abstract}


\section{Introduction}
\IEEEPARstart{R}{e}cent years have seen notable advancements in the evolution of Advanced Driver Assistance Systems (ADAS) and Autonomous Vehicles (AVs). 
The overarching goal of this research is to establish a service system that prioritizes safety and comfort for humanity~\cite{adas}. 
However, there remains room for enhancement in ensuring driving safety.
From 2014 to September 20, 2023, the DMV received 655 Autonomous Vehicle Collision Reports~\cite{california-dmv}. 
According to the World Health Organization (WHO), the annual global death toll due to road accidents stands at a staggering 1.3 million. 
More than half of all road traffic deaths and injuries involve vulnerable road users, such as pedestrians, cyclists, motorcyclists, and their passengers~\cite{who,niu2022auditory}. 
Therefore, developing a system to predict important objects in driving scenarios is crucial to advance ADAS and AV technology.

\begin{figure}
  \centering
  \includegraphics[width=0.9\linewidth]{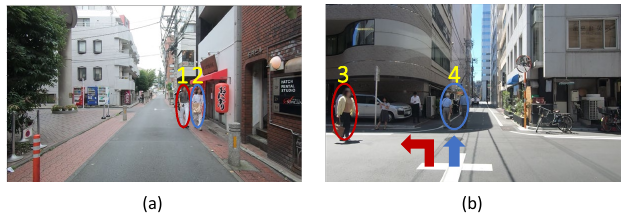}
  \caption{\textbf{Overview of our inspiration.} As shown in the figure (a), which is easily understandable for human drivers, the significance of the first pedestrian outweighs that of the second, as humans recognize that pedestrian 1 is is walking within the driving lane, whereas pedestrian 2 is not.
    Human drivers grasp a inherent relationship between pedestrians and line markings. 
  As depicted in Figure (b), even within the same intersection, the significance of objects can vary based on fluctuations in the driver's intention. 
  If the driver intends to turn left, pedestrian 3 becomes the most important, whereas if the intention is to go straight, pedestrian 4 takes precedence.}
  \label{fig:first}
\end{figure}

Driving scenes Important Object Detection (DIOD) involves assessing all participants in a driving scenario to predict objects that could significantly influence future control decisions. 
The potential applications of DIOD are wide-ranging. 
It can be seamlessly integrated into ADAS to help drivers identify hazardous objects and mitigate potential dangers. 
Additionally, DIOD systems can support AVs in detecting potential hazards, streamlining subsequent tasks like trajectory prediction, decision-making, and motion planning.

Currently, a considerable body of research~\cite{drive,salient,driverattention,hammerdrive,whatmakes,karim2022attention,zhang2020interaction} simplifies DIOD as a mere object detection task. 
This approach treats all participants as independent entities and directly predicts the importance of each object to the ego-vehicle.
They frequently neglect the intricate relationships among all participants, whereas human drivers instinctively consider these interrelations when predicting the importance of objects.
This kind of methods fails to distinguish the importance difference illustrated in Figure \ref{fig:first}. 
\IEEEpubidadjcol
As shown in the figure \ref{fig:first}(a), which is easily understandable for human drivers, the significance of the first pedestrian outweighs that of the second, as humans recognize that pedestrian 1 is is walking within the driving lane, whereas pedestrian 2 is not.
Human drivers grasp a inherent relationship between pedestrians and line markings. 
In actual driving situations, human drivers take into account not only the influence of individual objects on the ego-vehicle but also the interactions between these objects.

Consequently, comprehending the interactions among participants in driving scenes holds paramount importance for DIOD.
There are existing methods~\cite{kim2021hotr, tian2020road, yu2021scene, peng2023parallel, zhang2020interaction, zhang2022efficient, zou2021end} designed to instruct models about relationships within scenes.
However, a common limitation among these methods is their reliance on manually annotated relationship labels to guide machine learning models. 
Unfortunately, manual annotation falls short in capturing all object relationships. Some methods focus solely on annotating spatial relationships within a single scene, while others concentrate on the relationships of specific participants, neglecting many crucial participants.
Moreover, manually defined relationships are crafted for human comprehension but may not necessarily be optimal for machine understanding. 
Hence, the introduction of a "driving scene relationship self-understanding" module becomes crucial for enabling the model to autonomously learn interrelationships among all objects.
 
While previous research has primarily focused on enhancing the extraction of visual information from driving scenes to improve DIOD accuracy, it's important to acknowledge that the significance of key objects can shift in real driving situations, influenced by the driver's intentions.
As depicted in Figure \ref{fig:first}(b), even within the same intersection, the significance of objects can vary based on fluctuations in the driver's intention. 
If the driver intends to turn left, pedestrian 3 becomes the most important, whereas if the intention is to go straight, pedestrian 4 takes precedence.
Hence, predicting important objects should consider not only the driving scene but also the driver's intention.

\begin{figure}
  \centering
  \includegraphics[width=1.0\linewidth]{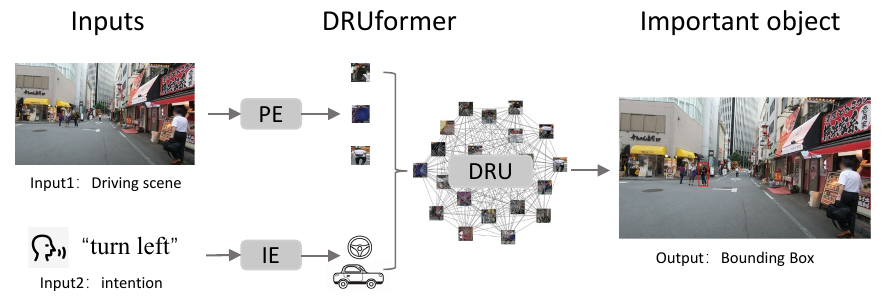}
  \caption{\textbf{Overview of our pipeline.} Our DRUformer model takes both the "driving scene" and the driving "intention" as input, to ascertain the most "important object" within the scene while considering the relationships among all the participants.}
  \label{fig:overview}
\end{figure}

Another critical concern is that current DIOD models heavily depend on CNN architectures. However, for AV tasks, Hu et al. \cite{hu2023planning} have illustrated that varying different modal information across tasks may result in low information exchange efficiency during CNN feature extraction. 
Hence, it is essential to propose a DIOD model based on an architecture that distinguishes itself from CNNs. 
Within this architecture, the DIOD model should adeptly blend information from both the driving scene and driving intention. 
Moreover, it must seamlessly transmit this combined information to the subsequent module for relationship understanding without any data loss.

To address the aforementioned issues, we introduced a model named \textbf{D}riving scene \textbf{R}elationship self-\textbf{U}nderstanding trans\textbf{former} (DRUformer) for the DIOD task, as depicted in Figure \ref{fig:overview}. 
Our model is designed to simultaneously integrate both the driving scene (driving frame) and driving intention (textual intention command) while taking into account the relationships between participants, ultimately enabling the prediction of important objects within the driving scene.
Our model primarily comprises three modules: 
the Participants Extractor Module (PE), which extracts information about all participants in the driving scene; 
the Intention Extractor Module (IE), responsible for extracting driving commands related to the driver's intention; 
and the Driving scene Relationship self-Understanding Module (DRU), which focuses on learning the interrelations between all objects without the mannual relationship definition.
The Transformer-based algorithms have been extensively proven as the most suitable framework for multi-modal tasks at the current stage~\cite{radford2021learning, li2022blip, yu2022coca,bao2022vlmo,xu2022groupvit,zhou2022conditional,10186652,niu2023r}. 
Therefore the entire approach is constructed upon the Transformer framework, facilitating information exchange optimization between all modules and enhancing downstream tasks.
To evaluate the effectiveness of our proposed model, we conducted comprehensive qualitative and quantitative analyses on the largest existing important object detection dataset, the DRAMA~\cite{malla2023drama} dataset.

In summary, our contributions are as follows:
\begin{itemize}
\item We propose a novel model that predicts important objects by taking into account the relationships between all driving participants as well as the driver's driving intention.
\item We introduce a driving scene relationship self-understanding module to learning the relationships between the participants without the need for manual relationship definition.
\item We conducted extensive quantitative and qualitative analyses on the DRAMA dataset, and our model significantly outperforms the baseline, resulting in substantial improvements in the DIOD task.
\end{itemize}

The subsequent chapters will be organized as follows:
In the section~\ref{sec:related work}, we will provide an overview of current developments related to the DIOD task and research related to relationship understanding.
The section~\ref{sec:method} will outline our DRUformer model and the construction of its individual modules.
In the section~\ref{sec:experiment}, we will detail the datasets employed in our experiments, the experimental design, and the results achieved.
The section~\ref{sec:conclusion} will offer a summary of the contributions made throughout our paper.

\section{Related work\label{sec:related work}}
\subsection{Risk object detection}
The task of risk object detection can primarily be categorized into two approaches: direct detection and indirect detection.

Direct detection aims to use the powerful regression capabilities of neural networks to mimic human-annotated datasets of important objects based on supervised tasks. 
In essence, these methods classify all objects into two categories: important and non-important. 
Several studies~\cite{drive,salient,driverattention,hammerdrive,umedamodeling} focus on utilizing human gaze (attention) information from experienced drivers as the supervised training labels. 
This label information is used to predict pixel-level attention regions through neural networks. 
However, this approach encounters two significant challenges. 
Firstly, human drivers may become distracted during driving, leading to their gaze fixating on objects unrelated to the driving task, such as interesting billboards. 
Secondly, the pixel-level information may not always effectively clustered around objects. 
Therefore, research~\cite{whatmakes,karim2022attention,zhang2020interaction} have adopted the use of object-level bounding box information as the learning target. 
This shift in focus enhances the prediction of risk objects, ensuring that it primarily concentrates on objects relevant to driving.

Indirect detection methods indirectly predict risk objects or risk regions through proxy tasks such as brake prediction and steering wheel angle prediction. 
Research~\cite{zhang2020interaction,wang2019deep,li2020make} involves pre-training a model for driving behavior prediction and subsequently identifying hazardous objects by removing extraneous objects from the scene. 
The truly risky object is believed to have a significant impact on driving outputs. 
Research~\cite{kim2019grounding,kim2017interpretable} focuses on training an end-to-end model for image-to-steering angle prediction and conducts causality analysis using attention maps to forecast hazardous pixel areas within the CNN model. 
It's worth noting that indirect prediction methods are primarily intended for offline use and are particularly well-suited for explainability AI but not optimal for online DIOD tasks.

While the previously mentioned methods have undeniably advanced the field of DIOD, they still grapple with the following issues:
\begin{itemize}

\item They often overlook the intricate relationships among all participants, whereas human drivers instinctively consider these interrelations when predicting the importance of objects.

\item These methods do not take into account the importance of human driving intentions, especially in advanced autonomous vehicles.

\item Most of their models rely on CNN architectures, which may not be ideal for conducting multi-modal research and facilitating information exchange for downstream tasks.
\end{itemize}

To address these challenges, we propose a transformer-based multi-modal model that takes into account both the interrelationships between the participants as well as the driver's driving intention.

\subsection{Relationship understanding}
The Relationship Understanding (RU) task involves comprehending the relationships between objects within a scene through various methodologies. 
Currently, RU plays a pivotal role in tasks such as Human-Object Interaction (HOI) and Scene-Graph Generation (SG)~\cite{kim2021hotr,zhang2022exploring,zhang2023exploring, tian2020road, yu2021scene, peng2023parallel, zhang2020interaction, zhang2022efficient, zou2021end,liao2022gen,wang20231st,kumar2021scene}. 
RU can be broadly classified into two approaches: two-stage (sequential) and one-stage (parallel) methodologies.

Two-stage methods typically encompass entity detection and relationship classification as two sequential steps. 
In the initial stage, an off-the-shelf object detector is employed to identify all objects, while in the subsequent stage, the detected objects are paired, and relationships between these paired objects are predicted. 
These frameworks aim to optimize object detection separately, either to enhance object detection accuracy or to improve HOI tasks through interaction classification optimization. 
Nonetheless, the primary challenge with this pipeline lies in the fact that the two stages cannot be concurrently fine-tuned, often resulting in decreased detection accuracy and efficiency.

Conversely, one-stage methods treat HOI as a Set Prediction problem, simultaneously predicting object pairs boxes and relationships between objects. This approach offers a more direct and efficient methodology with reduced time complexity~\cite{zhang2021mining}. However, one-stage methods frequently require intricate post-processing steps.

In the realm of autonomous driving, Yu et al.~\cite{yu2021scene} have employed a two-stage SG method to define spatial location relationships in highway driving situations, transforming the driving scene into a driving scene-graph. 
This approach is limited to specific scenarios and only addresses spatial location relationships (e.g., left, front, right). 
Another study by Tian et al.~\cite{tian2020road} also employs a two-stage SG method for the driving SG task, encompassing a broader range of scenarios, including pedestrians, bicycles, and more agents. 
However, this method still falls short in accounting for the intricate relationships between the driving vehicle, traffic signals, and driving direction.

Despite the significant advancements brought about by the aforementioned methods in the DRU task, they still confront the following challenges:
\begin{itemize}

\item The relationships defined within their scenes often rely on manual definitions, which prove insufficient for the complexity of the driving environment, where not all relationships can be predefined.

\item The definitions of relationships between objects predominantly focus on spatial relationships, whereas relationships can be more multifaceted. For instance, the connection between a pedestrian and a traffic signal exists even if they are spatially distant.

\item Existing driving RU methods primarily adhere to a two-stage approach, resulting in relationships that are often constrained by the outcomes of object detection. 
Simultaneously optimizing relationship generation and object detection remains a challenge in this context.
\end{itemize}

To tackle these challenges, we introduce a DRU module that empowers our DRUformer to learn relationships between all participants without relying on human-defined labels. 
To concurrently enhance participant detection and relationship comprehension, we treat each participant as a transformer token, allowing for direct integration into the DRU module.

\section{Method\label{sec:method}}
\begin{figure*}
  \centering
  \includegraphics[width=1.0\linewidth]{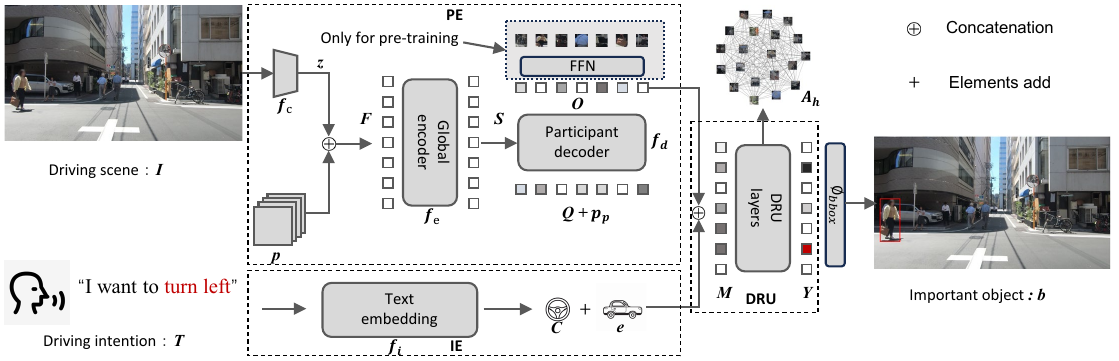}
  \caption{\textbf{Architecture of the DRUformer}. This is the pipeline of our DRUformer model. We feed the driving scene $\bm{I}$ and the driving intention $\bm{T}$ into our DRUformer and obtain the important object $\bm{b}$. In this context, the "PE" module signifies the participants extractor, the "IE" module signifies the driving intention extractor, and the "DRU" module represents the driving relationship self-understanding module.}
  \label{fig:pipeline}
\end{figure*}

To leverage the success of the transformer-based multimodal approach, we present a transformer-based multimodal model designed for the detection of critical objects in driving scenes, which we have named \textit{DRUformer}. 
The model's architecture is depicted in Figure \ref{fig:pipeline}. 
The DRUformer model comprises three key modules:
\begin{itemize}
    \item The \textit{Participants Extractor} module (PE) is tasked with capturing both the positional and semantic information of objects that are relevant to the act of driving.

    \item The \textit{Intention Extractor} module (IE) is specially designed to collect the driving intention from the driver intention command.

    \item The \textit{Driving relationship self-understanding} module (DRU) is dedicated to comprehending the interactions that transpire among all participants within the driving scene without the need for manual definition.
\end{itemize}

\subsection{Participants Extractor Module}
This module is designed to extract the participants in the driving scene.

The PE module is designed based on the powerful transformer-based object detection model called DETR~\cite{carion2020end}. 
PE comprises a standard CNN backbone referred to as $f_c$, a standard transformer encoder denoted as $f_e$, a standard transformer decoder designated as $f_d$ , and a simple feed forward network (FFN) to make the participant bounding box prediction.
We should note that the FFN part is only utilized in the pre-training step and does not participate in the main training step.

For the driving scene $\bm{I}$ with size $H\times W \times 3$, we utilize $f_c$ to extract a lower-resolution feature map $\bm{z} \in \mathbb{R}^{H_S \times W_S \times D_{S}}$ encompassing the entire scene, $H_S \times W_S$ represents the size of the feature map and $D_{S}$ represents the dimension. 
In this phase, $\bm{I}$ is downsampled into global spatial features with dimensions $H_S \times W_S$, typical $H_S, W_S = \frac{H}{32} \frac{W}{32}$. 
The encoder $f_e$ expects a sequence as input, hence we flatten and embedding the spatial feature $\bm{z}$ into one dimension features $\bm{F} = \left\{ f_i \,|\, f_i \in \mathbb{R}^{D_d} \right\}_{i=1}^{H_S \times W_S}$, ${D_d}$ represents the channel dimension of the features, $D_d$ is often much smaller than $D_S$.
These features in $\bm{F}$ are then passed through $f_e$ for global feature extraction.

In the $f_e$ module, each $f_e$ layer consists of a multi-head self-attention (MHSA) module and a feed-forward network (FFN). 
Since the
transformer architecture is permutation-invariant, we integrate positional information from $\bm{F}$ by inputting the positional encoding $\bm{p} \in \mathbb{R}^{H_S \times W_S \times D_{S}}$ into the $f_e$ module. 
This results in global features  $\bm{S} = \left\{ m_i \,|\, m_i \in \mathbb{R}^{D_d} \right\}_{i=1}^{H_S \times W_S}$ that include position information.
These features are then used in subsequent decoder stages.

In the case of $f_d$, each $f_d$ layer includes a self-attention module, a cross-attention module, and an FFN. 
We initialize N queries denoted as $\bm{Q} = \left\{ q_i | q_i \in \mathbb{R}^{D_d} \right\}_{i=1}^{N}$, to extract object information from the global features $\bm{S}$. 
Since the decoder is also permutation-invariant, to incorporate positional information into these $\bm{Q}$ features, we include positional encoding $\bm{p}_{p} \in \mathbb{R}^{H_S \times W_S \times D_{S}}$. 
After processing through the $f_d$ module, we obtain $N$ scene participants tokens, denoted as $\bm{O} = \left\{ o_i | o_i \in \mathbb{R}^{D_d} \right\}_{i=1}^{N}$, which includes both the positional and semantic information of $N$ participants in the scene.
The $N$ scene participants tokens $\bm{O}$ are then independently decoded into box coordinates and class labels by
a FFN, resulting $N$ predictions.
As mentioned earlier, we aim to simultaneously optimize relationship generation and object detection. 
Therefore, we exclusively employ the FFN only during the pre-training phase. 
In the model training stage, we rely solely on scene participants tokens $\bm{O}$ for information propagation, omitting the use of FFN-generated boxes and class information.

The PE is calculated as follows:
\begin{align}
\bm{F} = \text{flatten}(f_c(\bm{I})), \\
\bm{S} = f_e(\bm{F} + \bm{p}), \\
\bm{O} = f_d(\bm{Q}+\bm{p}_p, \bm{S}).
\end{align}
where \text{flatten} represents the flatten operation.

\subsection{Intention Extractor Module}
This module is designed to extract the driver's driving intention command.

In our research, the "Intention Extractor" (IE) module, denoted as $f_i$, is constructed based on a text embedding model. 
The driver's text intention command, represented as $\bm{T}$, is tokenized by the IE module into an intention command token, denoted as $C \in \mathbb{R} ^{1 \times D_{d} }$. 
It's important to note that, for the convenience of subsequent relationship understanding part, $\bm{C}$ and $\bm{o_i}$ should have the same dimensionality.
The calculation formula is as follows:
\begin{equation}
\bm{C} = f_i(\bm{T}).
\end{equation}

\subsection{Driving Relationship Understanding Module}
To model dense interrelationships between the ego-vehicle and other participants, we employ a self-attention module to learn the mutual relationships between all objects. 

\begin{figure}
  \centering
  \includegraphics[width=0.6\linewidth]{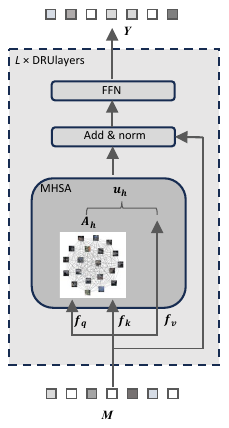}
  \caption{\textbf{Architecture of our DRU module.} $\bm{M}$ represents all the entities, $\bm{Y}$ represents all the entities with understanding the interrelationships. MHSA represents the multi-head self-attention module, and FFN represents the feed-forward network.}
  \label{fig:drulayer}
\end{figure}

Initially, we randomly initialize a learnable token to represent the ego-vehicle token, denoted as $\bm{e} \in \mathbb{R} ^{1 \times D_{d} }$, which serves as a carrier for the intention token $\bm{C}$, making it easier to learn.
Since we aim to learn the relationships among all objects in the scene, we then combine $\bm{C}$ and $\bm{e}$ together and concatenate them with the participants $\bm{O}$ to get all entities $\bm{M} \in \mathbb{R} ^{k \times D_{d} }$, where $k = N + 1$. This concatenation is expressed as:
\begin{equation}
\bm{M} = \text{norm}\left(\bm{C} + \bm{e}\right) \oplus \bm{O}.
\end{equation}
where $\text{norm} \left( \cdot \right)$ represents the layer normalization.

Subsequently, we feed all the entities in $\bm{M}$ into our DRU module to learn their relationships with each other. 
Our DRU module consists of $L$ layers of DRU layers. 
Figure \ref{fig:drulayer} visually illustrates the architecture of our DRU module.
Each DRU layer comprises MHSA and a FFN, with ${H}$ heads in the MHSA.
Each relationship head deals with the subspace $\bm{m} \in \mathbb{R} ^{k \times D_{h} }$, where $D_{h} = \frac{D_{d}}{H}$. 
The understanding mechanism is calculated as follows:
\begin{align}
\bm{A}_h &= \text{Softmax} \left( \dfrac{f_q\left( \bm{m} \right) f_k\left(\bm{m} \right) ^ T}{\sqrt{D_{h}}} \right), \\
&\bm{u}_h = \bm{A}_h \cdot f_v(\bm{m}).
\end{align}
where $f_q$, $f_k$, and $f_v$ represent the linear transformation layers. 
$\bm{A}_h \in \mathbb{R} ^{k \times k }$ is the entity relationship map for each heads.
It should be noted that in our research, the strength of the relationship between two objects is represented as the cosine similarity between the two objects, with a higher similarity indicating a stronger relationship.
Each heads output $\bm{u}_h \in \mathbb{R}^{ k \times D_h}$ is already understanding the relationship.

Finally, the output of each relationship head needs to be concatenated together and passed through an FFN to obtain the output of the DRU layer, denoted as  $\bm{Y} \in \mathbb{R} ^{k \times D_{d} }$, at which point $\bm{Y}$ has comprehended the relationships between all objects.

At this stage, the bounding boxes of the important objects $\bm{b} \in \mathbb{R}^{k \times 4}$ in the scene can be predicted using a predictor FFN $\phi_{bbox}(\cdot)$, as shown in the following formula:
\begin{equation}
    \bm{b} = \phi_{bbox} \left ( \bf{Y}  \right ). 
\end{equation}

\subsection{Training}
Following the set-based training process of DETR, we first match each ground truth important object bounding box with its best-matching prediction by the bipartite matching with the Hungarian matching algorithm\cite{carion2020end}.
Then the loss is produced between the matched predictions and the corresponding ground truths for the final back-propagation. 

Similar to DETR, the loss of DRUformer is composed of 3 parts: the box regression loss $\mathcal{L}_{b}$, the intersection-over-union loss $\mathcal{L}_{GIoU}$\cite{Giou}, and the object important class loss $\mathcal{L}_{c}$.
Loss function is shown as follows: 
\begin{equation}
    \mathcal{L} = \lambda_{b} \mathcal{L}_{b} + \lambda_{GIoU} \mathcal{L}_{GIoU} + \lambda_{c} \mathcal{L}_{c}.
\end{equation}
where $\lambda _{b}$, $\lambda _{GIoU}$ and $\lambda _{c}$ are the hyper-parameters for adjusting the weights of each loss.

\section{Experiment\label{sec:experiment}}
The code of our paper and more detection results will be here \url{https://github.com/oniu-uin0/DRU_former}.
\subsection{Experiment Setting}
\subsubsection{Dataset}
To accurately evaluate the capabilities of our proposed model, we chose to conduct tests on the largest existing important object evaluation dataset, DRAMA. 
This dataset was collected by HONDA Corporation in the Tokyo region of Japan and comprises a total of 17,785 scenarios. 
These scenarios encompass three main types of environments: wider roads (51.5\%), intersections (32.6\%), and narrow str. ets (15.9\%). 
This diversity in environments allows us to thoroughly evaluate our model's important object detection capabilities under various road conditions.
The majority of scenarios (17,066, 95.95\%) are labeled with the driving direction of the ego-vehicle in that scenario and the bounding box information of important object. 
The driving direction serves as the driving intention of our model. 
The important objects in the dataset include vehicles (71.9\%), pedestrians or cyclists (19.6\%), and infrastructure (8.5\%). 
This composition enables us to test the effectiveness of our model in detecting important objects of different sizes.

To ensure a robust evaluation, we shuffled the dataset and then divided it into training, validation, and test sets in a 70:15:15 ratio for training and testing purposes.
\subsubsection{Evaluation metrics}
In our research, we utilized two standard evaluation metrics: Accuracy (ACC) and Mean Intersection over Union (mIoU).

mIoU assesses the mean overlap between the predicted bounding boxes, denoted as $y_{pred}$, and the ground truth bounding boxes, represented as $y_{label}$. 
A higher mIoU value means more precise predictions of the bounding boxes. 
The calculation formula for mIoU is as follows:

\begin{equation}
    mIoU = \frac{1}{N} \sum_{i=1}^{N} \frac{I(y_{pred}, y_{label})}{U(y_{pred}, y_{label})}
\end{equation}
where $N$ represents the number of samples, and$I(y_{pred}, y_{label})$ and $U(y_{pred}, y_{label})$ represents the Intersection and Union between $y_{pred}$ and $y_{label}$ for each sample, respectively.

ACC evaluates whether our model correctly identifies the presence of important objects in the scene and accurately deter mines their locations within the image. 
A higher ACC indicates  a more effective ability to predice the presence of important objects. 
The calculation formula for ACC is as follows:

\begin{equation}
    ACC = \frac{1}{N} \sum_{i=1}^{N} [y_{pred} = y_{label}],
\end{equation}
where N represents the number of samples, and $[y_{pred} = y_{label}]$ equals 1 if the $ \frac{I(y_{pred}, y_{label})}{U(y_{pred}, y_{label})} > 0.5$.

\subsubsection{Implementation Details}
All of our experiments were conducted on a server with 8 NVIDIA RTX A6000 GPUs. During the experiment process, the hyperparameters were set as follows:
For the CNN part, we utilized the standard ResNet50 architecture and a 6-layer standard Transformer encoder to extract global features.
In the decoder part, we used a 6-layer architecture to extract participants in driving scenes.
In the DRU module's relationship attention section, we employed an MHSA with 8 heads for relationship understanding.
The initial learning rate was set to 0.0001, and the weight decaying rate was $10^{-4}$ per 200 epochs.
We applied scale augmentation similar to DETR, resizing the input images such that the shortest side was at least 480 pixels and at most 800 pixels, while the longest side was at most 1333 pixels.
The size of $L_{DRU}$ was set to 3 layers.
During the training phase, the batch size for each epoch was set to 8, and we trained for 400 epochs to obtain our experimental weights.

\subsection{Quantitative analysis}
Table \ref{tab:SOTA} presents the position results of our method and other SOTA baselines on the DRAMA dataset for important object detection.
In this table, \textit{ICL} represents independent captioning and localization, \textit{LCP} represents localization with captioning prior, \textit{OF} represents the Optical Flow for the corresponding frames.
It is evident that our method, when provided with only the driving frames as input, exhibits a substantial improvement of 12.3\% in ACC and 16.2\% in mIoU compared to LCP with optical flow. 
Notably, LCP not only incorporates optical flow but also leverages scene captions for detection assistance, yet it still falls significantly behind our DRU-equipped approach.

In comparison to LCP without optical flow assistance, our method achieves a 14\% increase in ACC and a 17.9\% boost in mIoU. 
We also compared our method to SOTA approaches for general object detection tasks such as Mask R-CNN\cite{he2017mask} and DETR\cite{carion2020end}, and our method exhibited significant improvements over both of these methods. 
This convincingly demonstrates that our DRUformer delivers outstanding performance in the DIOD task on the DRAMA dataset, even without optical flow, showcasing its superior capabilities.
\begin{table}[!t]
\caption{DIOD performances comparing with the SOTA methods on the DRAMA dataset 
\label{tab:SOTA}}
\centering
\begin{tabular}{|c||c|c|}
\hline
Method & mIoU & Acc ($IoU>0.5$) \\
\hline
ICL without OF\cite{malla2023drama} & 0.553 & 0.617 \\
ICL\cite{malla2023drama} & 0.533 & 0.593 \\
LCP without OF\cite{malla2023drama} & 0.597 & 0.667 \\
LCP\cite{malla2023drama} & 0.614 & 0.684 \\
Mask R-CNN\cite{he2017mask} & 0.589 & 0.668 \\
DETR\cite{carion2020end} & 0.644 & 0.671 \\
DRUformer & \bf{0.776(16.2$\%\uparrow$)} & \bf{0.807(12.3$\%\uparrow$)} \\
\hline
\end{tabular}
\end{table}

\subsection{Qualitative analysis}
\begin{figure*}
  \centering
  \includegraphics[width=1.0\linewidth]{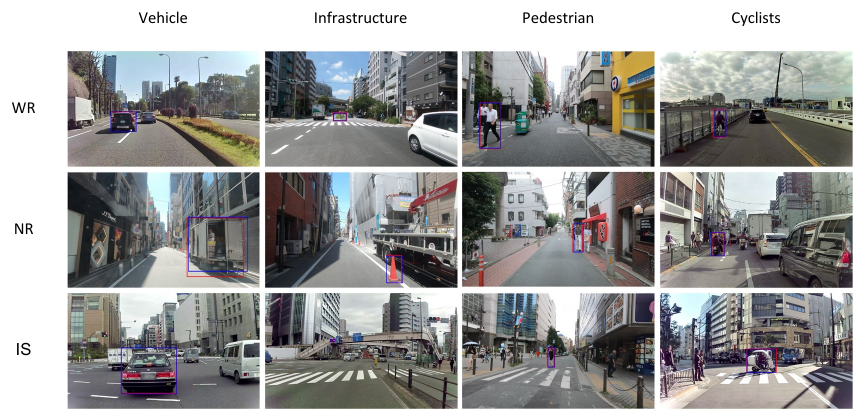}
  \caption{\textbf{Qualitative analysis of our DRUformer}. These are our model's important object detection results for various scenes and different types of objects. WR represents the Wider Road scene, NR represents the Narrow Road scene, and IS represents the intersection scene. The blue boxes represent the detection results from our model, while the red boxes represent the labels in the dataset.}
  \label{fig:scene}
\end{figure*}

Figure \ref{fig:scene} presents the position results of our algorithm in various road scenarios, including Wide Road (WR), Narrow Road (NR), and Intersection (IS), with important objects such as vehicles, pedestrians, cyclists, and infrastructure.
This visualization allows us to intuitively observe that our method, when predicting important objects, considers not only positional relationships but also the interactions between objects. 
For example, in the NR scenario pedestrian column, there are two pedestrians walking side by side on the right side.
Despite they close to each other, our model understands that only the pedestrian inside the blue box is within the lane (considering the relationship between the pedestrian and the lane). 
Therefore, the pedestrian inside the blue box is deemed the most important object, rather than both of the pedestrians.
Furthermore, our method exhibits excellent detection performance for objects of different sizes. 
For instance, in the IS scenario, vehicles are larger objects, while infrastructure objects like traffic signals are smaller. 
Figure \ref{fig:scene} also illustrates that our method can predict not only dangerous dynamic objects but also static traffic signals.
It is evident that our approach performs well in different driving scenarios and different important objects, providing robust object detection results.

\subsection{Ablation studies}
We conducted three ablation experiments to validate the effectiveness of the proposed modules.
\subsubsection{With Intention vs. Without Intention}
Table \ref{tab:intention} presents the localization results of our DRUformer in two scenarios: with the driving Intention (IN) and without IN. 
It is clear that when considering IN, our model shows a significant improvement in the DIOD (Driving Intention-Oriented Object Detection) task, with a 3.4\% increase in ACC and a 3.6\% increase in mIoU compared to the scenario without DIC. 
This demonstrates the effectiveness of the Driving Intention Command.

\begin{table}[!t]
\caption{Ablation results for the impact of the driving intention\label{tab:intention}}
\centering
\begin{tabular}{|c||c|c|}
\hline
Method & mIoU & Acc ($IoU>0.5$) \\
\hline
DRUformer without IN & 0.742 & 0.771 \\
DRUformer with IN & \bf{0.776(3.4$\%\uparrow$)} & \bf{0.807(3.6$\%\uparrow$)} \\
\hline
\end{tabular}
\end{table}

\begin{figure}
  \centering
  \includegraphics[width=1.0\linewidth]{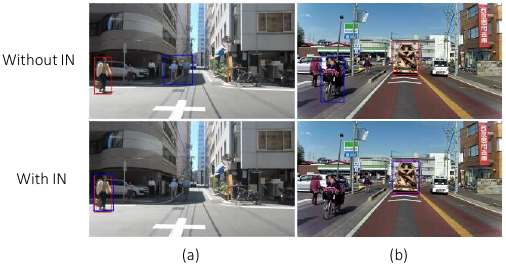}
  \caption{\textbf{Visual ablation results for driving intention.} IN represents the intention. The first column shows the DIOD results of our method when excluding the driving intention module, while the second column displays the detection results of our model including the driving intention.}
  \label{fig:intention}
\end{figure}

Figure \ref{fig:intention} provides a more intuitive demonstration of the impact of driving intention on our model. 
It can be observed that without the IN, our model may predict important objects in different directions. 
For example, in the scenario depicted in Image 3 (a), the model without IN predicts that the two individuals directly in front of the vehicle are important objects, while the actual label indicates the person in the red box on the left is important. 
However, in this case, it doesn't necessarily imply that the model's prediction is incorrect. 
If our vehicle needs to move directly forward, then the objects directly in front are indeed the most important.
After providing the IN for a left turn, the model can correctly predict the important objects.
It is evident that IN is helpful for detecting important objects in scenarios where there is direction ambiguity.

\subsubsection{With DRU module vs. Without DRU module}
Table \ref{tab:DRU} presents the localization results of our DRUformer model in two scenarios: with the DRU module and without it. 
It is evident that when considering inter-object relationships, our model demonstrates a significant improvement in the DIOD task compared to the scenario without the DRU module. 
Specifically, there is a 12.5\% increase in ACC and an 11.7\% increase in mIoU, providing clear evidence of the effectiveness of our DRU module for the DIOD task.

\begin{figure*}
  \centering
  \includegraphics[width=0.9\linewidth]{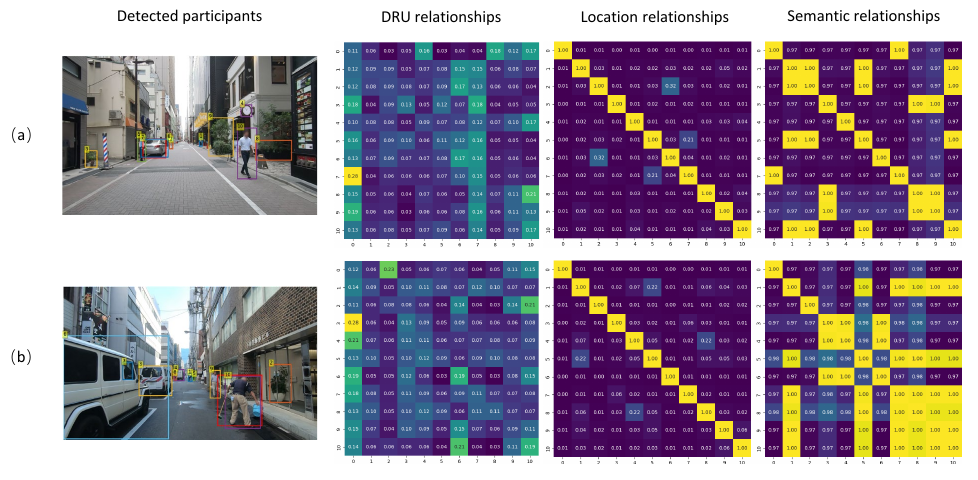}
  \caption{\textbf{Visual ablation results for the DRU module}. The first column display the results derived from the PE module after retaining the top 10 participants in the driving image. 
  The three columns on the right display the relationship networks acquired through our DRU module, the location relationship network based exclusively on bounding box information, and the semantic relationship network computed solely from object category information. 
  In the three relationship heatmaps on the right, "0" corresponds to the ego-vehicle, while "1-10" represent the 10 objects detected from the scene. 
  The complete heatmap illustrates the bidirectional relationship network among these 11 objects.}
  \label{fig:relation}
\end{figure*}

Figure \ref{fig:relation} visually illustrates the differences between the relationship network learned by our DRU module and other relationship networks based solely on manually defined relationships. 
We selected ten objects from all the objects detected by the PE module and combined them with the ego-vehicle to create the dense relationship map from our DRU module. 
Additionally, we generated location relationship networks and semantic relationship networks based on the bounding box information and class information of the detected objects. 
It is observable that the relationship network provided by our DRU module is more comprehensive and contains richer information. 
However, as mentioned earlier, manually defined relationships are convenient for human understanding but may not necessarily be suitable for machine learning. 
Our method, which allows the model to autonomously learn relationships and capture more nuanced relationship networks.

\begin{table}[!t]
\caption{Ablation results for the impact of the DRU\label{tab:DRU}}
\centering
\begin{tabular}{|c||c|c|}
\hline
Method & mIoU & Acc ($IoU>0.5$) \\
\hline
DRUformer without DRU & 0.659 & 0.682 \\
DRUformer & \bf{0.776(11.7$\%\uparrow$)} & \bf{0.807(12.5$\%\uparrow$)} \\
\hline
\end{tabular}
\end{table}

\subsubsection{Performance with different DRU layers}
Table \ref{tab:druformer_layers} showcases the influence of varying the number of DRU layers in our DRUformer on the DIOD task. 
We assessed the model's performance with 1 to 6 layers of DRU and monitored changes in Acc and mIoU. 
Notably, the optimal performance is attained with three layers. 
However, as the number of layers increases, the detection performance starts to deteriorate.

\begin{table}[!t]
\caption{DRUformer Performance with Different DRU Layers\label{tab:druformer_layers}}
\centering
\begin{tabular}{|c||c|c|c|}
\hline
Method & mIoU & Acc ($IoU>0.5$) & DRU layers \\
\hline
DRUformer & 0.694 & 0.717 & 1 \\
DRUformer & 0.712 & 0.749 & 2 \\
DRUformer & \bf{0.776} & \bf{0.807} & 3 \\
DRUformer & 0.748 & 0.776 & 4 \\
DRUformer & 0.733 & 0.758 & 5 \\
DRUformer & 0.686 & 0.745 & 6 \\
\hline
\end{tabular}
\end{table}

\section{discussion}
Although our DRUformer model achieved excellent results in the DIOD task, we did not conduct the following experiments due to limitations in graphics card memory. In our experiments, the driving scene section only selected images from the last key frame of the driving video as the scene. However, in this scenario with a single image, we did not consider temporal information, focusing solely on spatial information. Even in the case of a single image, our model already occupied 45GB of memory, while the A6000 graphics card has only 48GB of memory. In the future, we plan to consider replacing the scene section with the entire video scene to enhance the temporal information of the PE module. Simultaneously, we intend to explore the addition of Optical Flow corresponding to the scene to enhance scene information.

\section{Conclusion\label{sec:conclusion}}
To the best of our knowledge, DRUfomer model stands out as the first multi-modal model for important object detection in driving scenes based on the transformer architecture. 
This innovative approach considers both the driving scene and the driver's intention. 
Additionally, we introduce a driving scene relationship self-understanding module, specifically designed for machine comprehension, to enhance important object detection. 
This module eliminates the need for manually defining interactive relationships between objects.
We conducted a comprehensive evaluation by comparing our method with other SOTA models on the largest important object detection dataset, DRAMA. 
The results of this comparison affirm the effectiveness of our model. 
Furthermore, we designed numerous ablation experiments to confirm the efficacy of our proposed intention extractor module and relationship self-understanding module.

While our research has successfully extracted driving intention commands from the driver, it currently encounters limitations in human-AV interaction. Effectively communicating with passengers or drivers is a critical area that needs improvement. In the next phase, our goal is to broaden our focus beyond the driver's intention commands to include more complex interactive commands.
In scenarios where the driver lacks a specific driving intention, we acknowledge the importance of driving intention and plan to adopt the path planning route as a representative driving intention in our research.

\section*{Acknowledgment}
This work was supported by Nagoya University, the Japan Society for the Promotion of Science (JSPS), and the Japan Science and Technology Agency (JST). JST Grant Number JPMJFS2120, JSPS KAKENHI Grant Number JP21H04892 and JP21K12073.

% {\appendix[Proof of the Hungarian algorithm]
% Use $\backslash${\tt{appendix}} if you have a single appendix:
% Do not use $\backslash${\tt{section}} anymore after $\backslash${\tt{appendix}}, only $\backslash${\tt{section*}}.
% If you have multiple appendixes use $\backslash${\tt{appendices}} then use $\backslash${\tt{section}} to start each appendix.
% You must declare a $\backslash${\tt{section}} before using any $\backslash${\tt{subsection}} or using $\backslash${\tt{label}} ($\backslash${\tt{appendices}} by itself
%  starts a section numbered zero.)}

\bibliographystyle{IEEEtran}

\bibliography{ref.bib}

\clearpage 

\section{Biography Section}
\begin{IEEEbiography}
[{\includegraphics[width=1in,height=1.25in,clip,keepaspectratio]{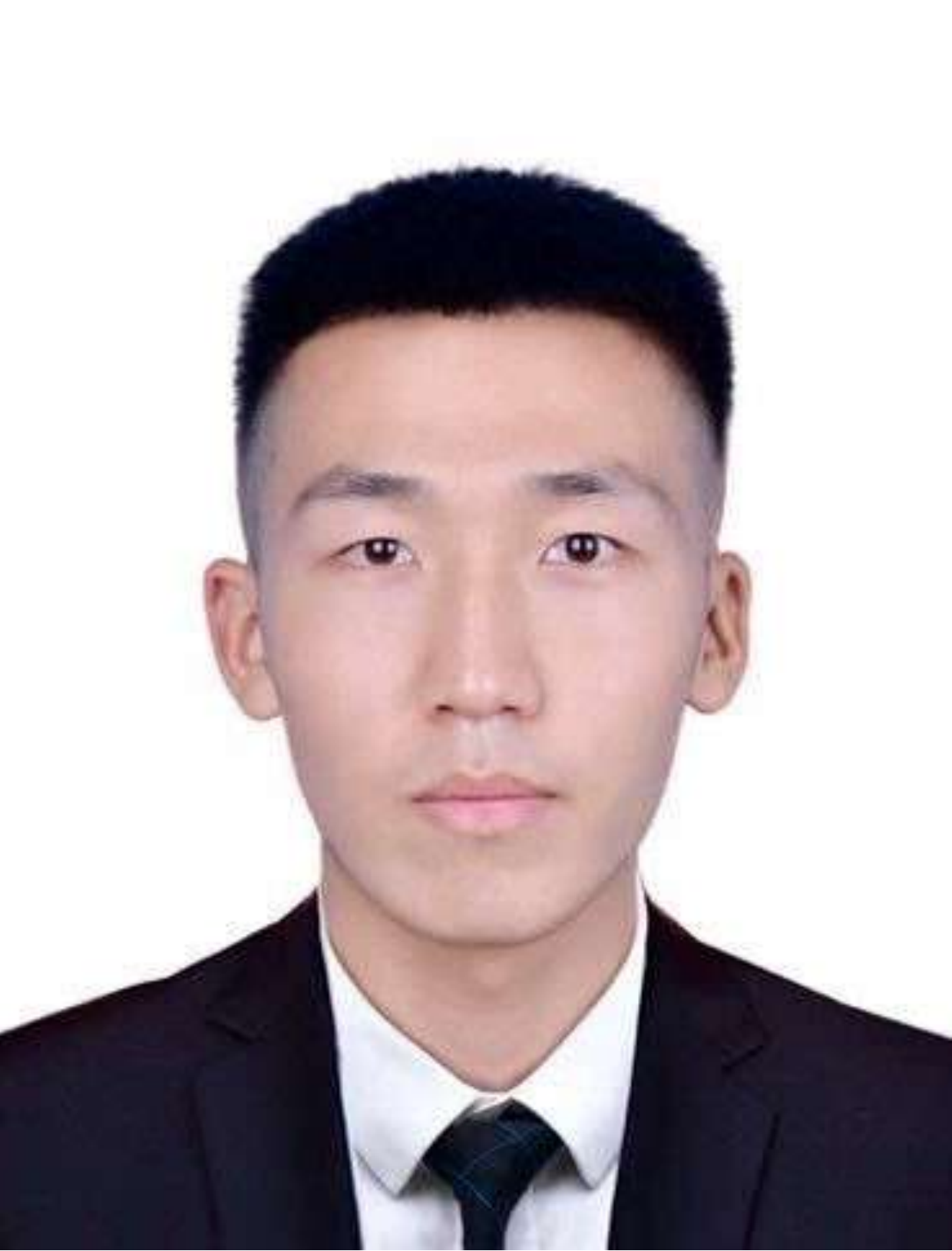}}]{Yingjie NIU}(Student Member, IEEE)
received the B.S. degree in Mechatronic Engineering from China Three Gorges University, Yichang, China, in 2018 
and the M.S. degree in Mechatronic Engineering from Southwest Jiaotong University, Chengdu, China, in 2021. 
He is currently working toward a Ph.D. degree in intelligent systems with the Graduate School of Informatics, Nagoya University, Nagoya, Japan.
His research interests include scene understanding, weakly supervised learning, and zero-shot learning.
\end{IEEEbiography}

\vspace{11pt}
\begin{IEEEbiography}
[{\includegraphics[width=1in,height=1.25in,clip,keepaspectratio]{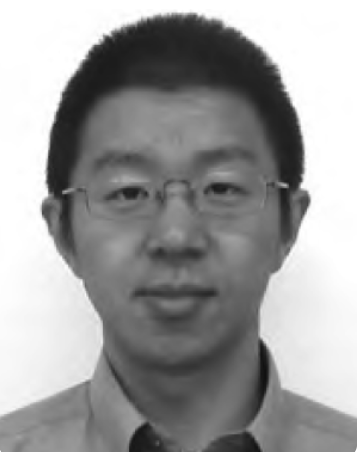}}]{Ming DING}
(Member, IEEE) received the M.S. and Ph.D. degrees in engineering from the Nara Institute of Science and Technology, Japan, in 2007 and 2010, respectively. In April 2010, he joined the Department of Mechanical Engineering, Tokyo University of Science, as a Postdoctoral Researcher. From October 2011 to February 2014, he was a Researcher with the RIKEN-TRI Collaboration Center for Human-Interactive Robot
Research, RIKEN. Since March 2014, he has been a Designated Assistant Professor with the Graduate School of Engineering, Nagoya University, Japan. Since May 2015, he has been an Assistant Professor with the Graduate School of Information Science, Nara Institute of Science and Technology. Since November 2019, he has been with the Institutes of Innovation for Future Society, Nagoya University, as a Designated Associate Professor. His current research interests include robot control,
human modeling, and human–machine interface. He is a member of JSR.
\end{IEEEbiography}
\vspace{11pt}

\begin{IEEEbiography}[{\includegraphics[width=1in,height=1.25in,clip,keepaspectratio]{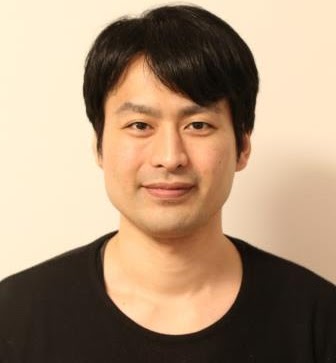}}]{Keisuke Fujii}(Member, IEEE) received his B.S., M.S., and Ph.D. degrees from Kyoto 
University in 2009, 2011 and 2014, respectively.
After his work as a postdoctoral fellow and research scientist at Nagoya University and RIKEN Center for Advanced Intelligence Project in Japan, he joined Nagoya University.
He is currently an associate professor at the Graduate School of Informatics.
His research interests are interdisciplinary studies among machine learning, behavioral sciences, and sports sciences. 
\end{IEEEbiography}

\vspace{11pt}
\begin{IEEEbiography}
[{\includegraphics[width=1in,height=1.25in,clip,keepaspectratio]{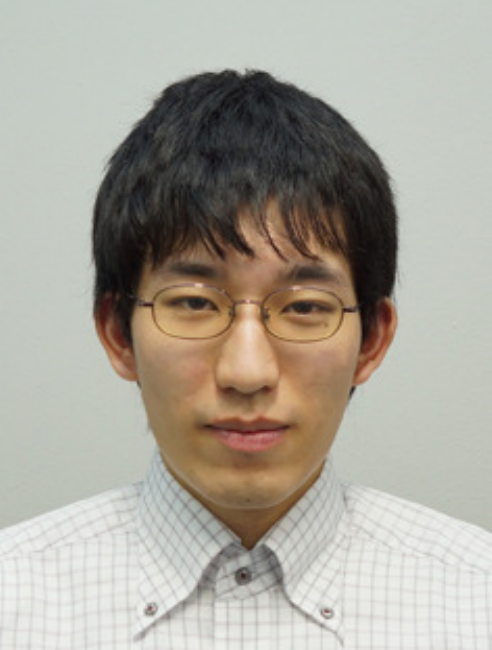}}]{Kento Ohtani} (Member, IEEE)
 received his B.E., M.E. and Ph.D. degrees from Nagoya University in 2013, 2015 and 2018, respectively. He is currently a Designated Assistant Professor at Nagoya University. His research interests are human behavior related signal processing include; spatial audio and autonomous driving.
\end{IEEEbiography}

\vspace{11pt}
\begin{IEEEbiography}
[{\includegraphics[width=1in,height=1.25in,clip,keepaspectratio]{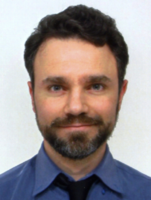}}]{Alexander Carballo} (Member, IEEE)
received the Dr. Eng. degree from the Intelligent Robot Laboratory, University of Tsukuba, Japan. From 1996 to 2006, he was a Lecturer with the School of Computer Engineering, Costa Rica Institute of Technology. From 2011 to 2017, he worked in LiDAR research and development at Hokuyo Automatic Company, Ltd. From 2017, he joined Nagoya University as Designated Associate Professor affiliated to the Institutes of Innovation for Future Society. Lastly, from 2022 he was appointed permanent Associate Professor at the Graduate School of Engineering in Gifu University, Japan. He is a professional member of IEEE Intelligent Transportation Systems Society (ITSS), IEEE Robotics and Automation Society (RAS), Robotics Society of Japan (RSJ), Asia Pacific Signal and Information Processing Association (APSIPA), the Society of Automotive Engineers of Japan (JSAE), and the Japan Society of Photogrammetry and Remote Sensing (JSPRS). His main research interests include LiDAR sensors, robotic perception, and autonomous driving.
\end{IEEEbiography}

\vspace{11pt}
\begin{IEEEbiography}
[{\includegraphics[width=1in,height=1.25in,clip,keepaspectratio]{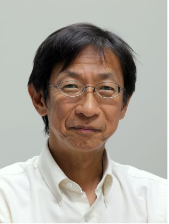}}]{Kazuya TAKEDA}
(Governors member, IEEE ITS Society; Governors member, APSIPA; Fellow, IEICE)
serves as a Vice President of Nagoya University and Professor at Nagoya University's Institute of Innovation for Future Society and Graduate School of Informatics. He is also a Director at Tier IV, Inc.
Dr. Takeda earned his Bachelor's, Master's, and Ph.D. from Nagoya University in 1983, 1985, and 1993, respectively. He has held positions at ATR (Advanced Telecommunication Research Laboratories) and KDD R\&D Lab, in addition to being a visiting scientist at MIT before rejoining Nagoya University in 1995.
From 2013 to 2022, Dr. Takeda was a Board of Governors member for both the IEEE ITS Society and the Asia-Pacific Signal and Information Processing Association (APSIPA). He chaired several scientific meetings, including FAST-zero 2017 and Universal Village 2016, and served as program chair for IEEE ICVES 2009 and IEEE ITSC 2017. Furthermore, he was the general chair of the IEEE Intelligent Vehicle Symposium (IV2021).
Dr. Takeda co-founded Tier IV, a university startup aimed at democratizing autonomous driving technologies through the development of the open-source software platform, Autoware.
His research primarily focuses on signal processing and machine learning of behavior signals and their applications. With over 150 journal papers, 9 co-authored/co-edited books, and 15 patents to his name, Dr. Takeda is a prolific contributor to his field. His achievements include the 2020 IEEE ITS Society Outstanding Research Award and six best paper awards from IEEE international conferences and workshops, in addition to various domestic awards.
\end{IEEEbiography}

\vfill

\end{document}